\documentclass[letterpaper]{article} 
\usepackage{aaai25}  
\usepackage{times}  
\usepackage{helvet}  
\usepackage{courier}  
\usepackage[hyphens]{url}  
\usepackage{graphicx} 
\usepackage{natbib}  
\usepackage{caption} 
\usepackage{colortbl}

\usepackage{algorithm}
\usepackage{algorithmic}

\usepackage{multirow}
\usepackage{booktabs}
\usepackage{subcaption}
\usepackage{amsmath,amsfonts,bm,amsthm,amssymb}
\usepackage{graphics,graphicx,caption,float,color}
\usepackage[capitalize]{cleveref}

\usepackage{appendix}

\usepackage[dvipsnames]{xcolor}

\usepackage{listings}
\usepackage{xspace}
 
\usepackage{hhline}
\usepackage{multicol}
\usepackage{booktabs} 
\usepackage{makecell} 
\crefname{section}{Sec.}{Secs.}
\Crefname{section}{Section}{Sections}
\Crefname{table}{Table}{Tables}
\crefname{table}{Tab.}{Tabs.}
\usepackage{newfloat}
\usepackage{listings}

\newcommand{\eg}{\emph{e.g.}}

\newcommand{\bs}{\boldsymbol}

\colorlet{best}{gray!60}
\colorlet{second}{gray!35}
\colorlet{third}{gray!10}

\DeclareCaptionStyle{ruled}{labelfont=normalfont,labelsep=colon,strut=off} 
\floatstyle{ruled}
\newfloat{listing}{tb}{lst}{}
\floatname{listing}{Listing}
\title{Efficient Gaussian Splatting for Monocular Dynamic Scene Rendering\\via Sparse Time-Variant Attribute Modeling}
\author{
    Hanyang Kong, Xingyi Yang, Xinchao Wang\thanks{Corresponding author.}
}
\affiliations{
    National University of Singapore

    hanyang.k@u.nus.edu, xyang@u.nus.edu, xinchao@nus.edu.sg
}

\usepackage{bibentry}

\begin{document}

\maketitle

\begin{abstract}
Rendering dynamic scenes from monocular videos is a crucial yet challenging task. The recent deformable Gaussian Splatting has emerged as a robust solution to represent real-world dynamic scenes. However, it often leads to heavily redundant Gaussians, attempting to fit every training view at various time steps, leading to slower rendering speeds. Additionally, the attributes of Gaussians in static areas are time-invariant, making it unnecessary to model every Gaussian, which can cause jittering in static regions. In practice, the primary bottleneck in rendering speed for dynamic scenes is the number of Gaussians. In response, we introduce Efficient Dynamic Gaussian Splatting (EDGS), which represents dynamic scenes via sparse time-variant attribute modeling. Our approach formulates dynamic scenes using a sparse anchor-grid representation, with the motion flow of dense Gaussians calculated via a classical kernel representation. Furthermore, we propose an unsupervised strategy to efficiently filter out anchors corresponding to static areas. Only anchors associated with deformable objects are input into MLPs to query time-variant attributes. Experiments on two real-world datasets demonstrate that our EDGS significantly improves the rendering speed with superior rendering quality compared to previous state-of-the-art methods.
\end{abstract}

\section{Introduction}

\begin{figure*}[htbp]
\centering 
\includegraphics[width=0.95\textwidth]{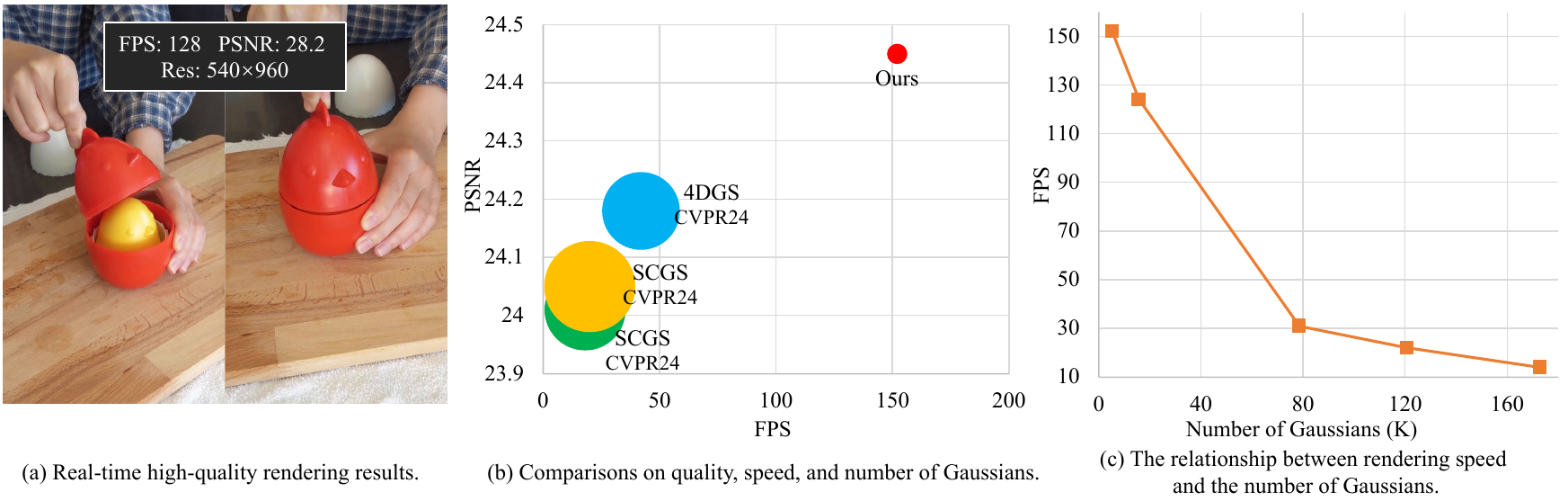}
\caption{(a) Given a set of monocular multi-view images and camera poses, our method achieves real-time rendering for dynamic scenes while maintaining high rendering quality. (b) Our method achieves promising rendering quality with faster rendering speed and fewer Gaussians. The radius of the circle is the number of time-variant Gaussians whose attributes need to be queried by MLPs. (c) The bottleneck of the rendering speed for dynamic scenes is the number of Gaussians. The fewer the number of Gaussians, the faster the rendering speed.} 
\label{fig:fig1}
\end{figure*}

Novel view synthesis (NVS) is a pivotal challenge in the field of 3D vision, essential for applications such as virtual reality, augmented reality, and film production. NVS involves generating images from arbitrary viewpoints or times within a scene, typically necessitating accurate reconstruction based on several 2D images. While recent advances in diffusion models~\cite{kong2025dreamdrone,yu2024wonderjourney,ma2024deepcache} have shown promise in NVS, dynamic scenes remain challenging due to the need to model complex motions and the need for real-time processing.

Recent advancements in 3D Gaussian Splatting (3DGS)~\cite{kerbl20233d} provide new tools to tackle these challenges by enabling real-time rendering. Extensions of 3DGS, such as \cite{huang2023scgs,wu20234dgs,yang2023deformable}, have further enhanced its ability to handle dynamic monocular scenes. For example, Deformable 3DGS~\cite{yang2023deformable} incorporates a deformable network to capture motion fields, enhancing 3DGS's adaptability to dynamic scenes. Similarly, 4DGS~\cite{wu20234dgs} utilizes a hexplane-based~\cite{hexplane} encoder to optimize deformation queries, making the process more efficient.

While previous methods deliver high-quality renderings, they tend to generate an excessive number of Gaussians, resulting in considerable redundancy. We analyze the relationship between the rendering speed \textit{w.r.t.} number of Gaussians based on~\cite{yang2023deformable} in \cref{fig:fig1}(c). As a result, the rendering speed declines as the number of redundant Gaussians increases. Inspired by recent work~\cite{lu2023scaffold} which improved efficiency in static scenes by sparsifying and reducing Gaussian points, we question: \textit{Can we achieve a more efficient and compact representation of dynamic scenes without compromising rendering quality?}

In this paper, we affirmatively answer this question by introducing Efficient Dynamic Gaussian Splatting (EDGS). This method efficiently represents dynamic scenes as a set of sparse time-variant attributes. At its heart, EDGS separately processes time-invariant attributes, time-variant attributes, and motions. This design allows precise control over each aspect of the scene's dynamics.

For time-invariant attributes, EDGS uses a sparse anchor-grid initialization to capture geometry and appearance. These attributes, such as color and opacity, are decoded by lightweight MLPs, enabling accurate deformation modeling.

Unlike prior methods that process all time-variant Gaussian attributes (e.g., location, scale, quaternion) indiscriminately through MLPs, EDGS adopts a selective approach. A tiny MLP filters static anchors, identifying those that are deformable. Only these deformable anchors are then processed by corresponding MLPs for their dynamic attributes. Notably, this selection process is trained in a fully unsupervised manner.

Moreover, we efficiently model the movements of deformable Gaussians in a sparse manner. Specifically, we track the movements of anchors linked to deformable objects at each time step. The offsets for corresponding Gaussians are computed using a radial basis function (RBF) kernel. This sparse approach enables precise and efficient rendering of dynamic changes.

By integrating these techniques, we carried out detailed experiments on two real-world datasets, NeRF-DS~\cite{yan2023nerfds} and HyperNeRF~\cite{park2021hypernerf}. We visualize the rendered novel views and show the rendering speed in \cref{fig:fig1}. Our method achieves a much faster rendering speed with a higher PSNR score and fewer points that the time-variant attributes need to be queried by MLPs. In summary, our contributions are summarized as follows:
\begin{itemize}
    \item We formulate deformable 3DGS through sparse, time-variant attribute modeling and introduce a novel unsupervised strategy to filter out Gaussians with time-invariant attributes. 
    \item We employ a classical kernel representation to formulate motion flow sparsely for the deformable Gaussians.
    \item Our EDGS achieves faster rendering speed compared with other state-of-the-art methods with higher rendering quality.
    
\end{itemize}

\section{Related Works}

\paragraph{Dynamic 3D scene reconstruction.}

3D Gaussian Splatting (3DGS)~\cite{kerbl20233d} offers a distinct representation for novel view synthesis with improved rendering quality and speed. Several concurrent works
~\cite{wang2024gflow,wu20234dgs,yang2023deformable,huang2023scgs} 
have adapted 3DGS for reconstructing dynamic scenes. For instance, Deformable 3DGS~\cite{yang2023deformable} formulates the dynamic field based on multi-layer perceptron (MLP) with differentiable rendering. 4DGS~\cite{wu20234dgs} improves the rendering speed with a compact network. SCGS~\cite{huang2023scgs} explicitly decomposes the motion and dynamic scenes into sparse control points and the deformation of Gaussians is controlled by its $k$-nearest control points. 

Though achieving promising rendered quality, the rendering speed for dynamic scenes is much slower than rendering static scenes (100 FPS \textit{v.s.} 20 FPS at 1K resolution). The bottleneck of the rendering speed for dynamic scenes is the number of Gaussian points. As shown in \cref{fig:fig1}, as more Gaussians are densified, an enormous amount of Gaussians are fed into the deformation network, which leads to slower rendering speed. Moreover, \cite{huang2023scgs,yang2023deformable,wu20234dgs} query the attributes of all Gaussians at each timestep, even though the static areas in real-world scenes are time-invariant and do not require querying by MLPs. As shown in \cref{fig:error}, the static areas of \cite{yang2023deformable,wu20234dgs}'s rendered results are still jittering across time. 

\paragraph{Grid-based rendering for acceleration.}

Grid-based representations are based on a dense uniform grid of voxels~\cite{fridovich2022plenoxels,kplanes,hexplane,lu2023scaffold} explicitly or implicitly. For instance, K-Planes~\cite{kplanes} applies neural planes to parameterize a 3D scene, with an additional temporal plane to accommodate dynamics. HexPlane~\cite{hexplane} further enhances the neural planes for time and space by factorizing space and time into compact neural representations. Scaffold-GS~\cite{lu2023scaffold} introduces anchor-voxel-based strategy to achieve reduced storage parameters for static scenes. Scaffold-GS can render a similar speed (100 FPS at 1K resolution) as 3DGS and the storage requirements are significantly reduced as Scaffold-GS only needs to store anchor points and MLPs for each scene. However, Scaffold-GS is designed for real-world static scenes and we are the first to adapt anchor-voxel-based strategy to dynamic scene reconstruction.

\section{Preliminaries}

In this section, we simply review the representation and rendering process of 3DGS~\cite{kerbl20233d} with the formula of deformable 3DGS~\cite{wu20234dgs, yang2023deformable} and our baseline method~\cite{lu2023scaffold}.

\begin{figure*}[htbp]
\centering 
\includegraphics[width=0.95\textwidth]{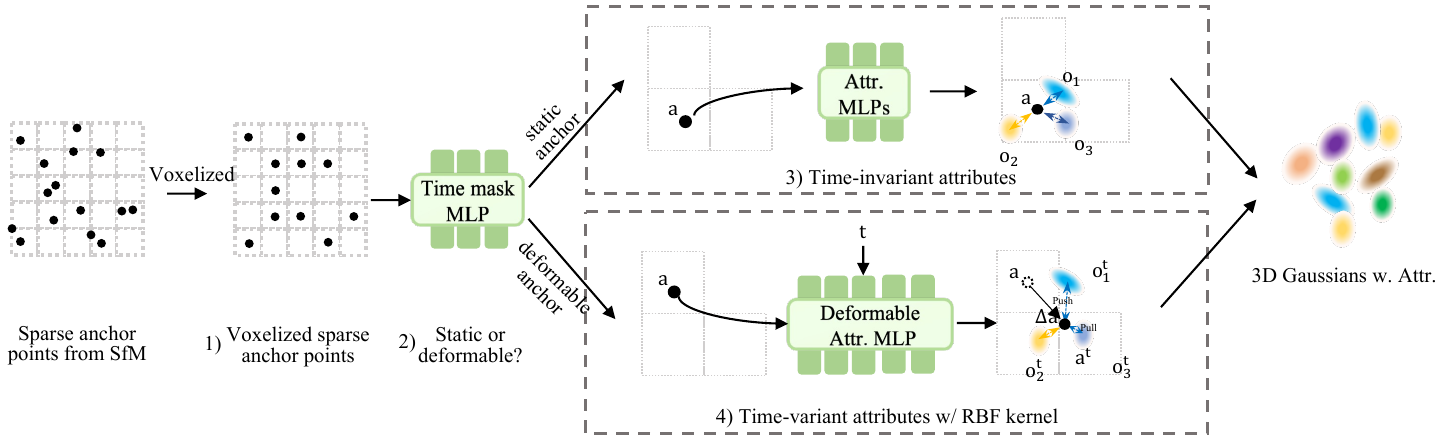}
\caption{\textbf{The pipeline of our EDGS.} 1) We first initialize voxelized sparse anchor points from Structure from Motion (SfM) points derived from COLMAP. 2) A time-mask MLP is applied to classify if the anchor belongs to the static area or the deformable area. 3) $k$ Gaussian offsets are initialized for each anchor $\boldsymbol{a}$ belonging to static area. The time-invariant attributes of each Gaussian, \textit{i.e.}, opacity, quaternion, scale, and color are calculated from its feature by corresponding tiny MLPs. 4) Time-variant attributes for anchors from dynamic areas are decoded by a deformable attribute MLP. RBF kernel function is employed to compute the location of each Gaussian at timestep $t$ by calculating the similarity between each Gaussian and its belonging anchor point. This pipeline is compact and efficient, featuring only a few tiny MLPs for the attributes of the Gaussians and a single network for deformations. Notably, the position of each anchor remains static and is not subject to updates.} 
\label{pipeline}
\end{figure*}

\subsection{Dynamic 3D Gaussian Splatting}

3DGS~\cite{kerbl20233d} is an explicit 3D representation in the form of point clouds. The process starts with 3D point clouds generated through Structure-from-Motion (SfM). Each Gaussian $G(\boldsymbol{x})$ is defined by a mean position $\boldsymbol{\mu}$ and a covariance matrix $\boldsymbol{\Sigma}$
\begin{equation}
G(\boldsymbol{x}) = e^{-\frac{1}{2} (\boldsymbol{x}-\boldsymbol{\mu})^T \boldsymbol{\Sigma}^{-1} (\boldsymbol{x}-\boldsymbol{\mu})},
\label{eq:gs}
\end{equation}
where $\boldsymbol{x}$ is a 3D point location within the 3D scene. $\boldsymbol{\Sigma}$ is formulated using a scaling matrix $\boldsymbol{S}$ and rotation matrix $\boldsymbol{R}$: 
\begin{equation}
\boldsymbol{\Sigma} = \boldsymbol{R}\boldsymbol{S}\boldsymbol{S}^T\boldsymbol{R}^T.
\label{eq:sigma}
\end{equation}

To render an image from a random viewpoint, 3D Gaussians are first splatted to 2D, and render the pixel value $\boldsymbol{C}$ by the following formula: 

\begin{equation}
\label{eq:splat}
    \boldsymbol{C} = \sum_{i\in N}c_i \alpha_i \prod_{j=1}^{i-1} (1-\alpha_i),
\end{equation}

where $c_i$ and $\alpha_i$ represent the density and view-dependent color of the point, and $N$ is the number of sorted Gaussians contributing to the rendering process.

To model the dynamic 3D Gaussians that vary over time, \cite{huang2023scgs,yang2023deformable,wu20234dgs} decouple the 3D Gaussians and the deformation field, separately. To be specific, given time $t$ and the location $x$ of 3D Gaussians as input, deformable 3DGS methods~\cite{huang2023scgs,yang2023deformable,wu20234dgs} apply the trained deformation MLP to produce offsets, which subsequently transform the 3D Gaussians from canonical space to the deformed space:
\begin{equation}
    \label{eq:deform}
    \Delta \boldsymbol{x}, \Delta \boldsymbol{r}, \Delta \boldsymbol{s} = \mathcal{F}(\gamma (\boldsymbol{x}, t)),
\end{equation}
where $\gamma$ is the encoding of space and time, $\Delta \boldsymbol{x}$, $\Delta \boldsymbol{r}$, and $\Delta \boldsymbol{s}$ are the offsets of the Gaussian location $\boldsymbol{x}$, quaternion $\boldsymbol{r}$, and scaling $\boldsymbol{s}$, respectively. Subsequently, the deformed 3D Gaussians $G(\boldsymbol{x} + \Delta \boldsymbol{x},\boldsymbol{r}+ \Delta \boldsymbol{r}, \boldsymbol{s} + \Delta \boldsymbol{s})$ are fed into the differential Gaussian rasterization pipeline for rendering novel views with various times $t$. All attributes of 3D Gaussians and the deformable network $\mathcal{F}$ are learnable and optimized end-to-end directly via training view reconstruction.

Though achieving promising results, the rendering speed of dynamic scenes remains significantly slower than that of static scenes (e.g., 20 FPS vs. 100 FPS for 1K resolution). This slowdown is due to the redundancy of Gaussians and the need to query the attributes of all Gaussians using the deformation network $\mathcal{F}$. A sparse Gaussian representation would accelerate the rendering process. Additionally, it is more efficient to query only the attributes of Gaussians associated with deformable objects, as the attributes of static areas remain unchanged over time.

\vspace{-0.1cm}
\subsection{Scaffold-GS} 

Scaffold-GS~\cite{lu2023scaffold} adheres to the framework of 3DGS for static scenes and introduces a storage-friendly anchor-based strategy. Scaffold-GS derives Gaussians from the anchors and deduces the attributes from the attributes of the attached anchors through MLPs, rather than storing Gaussians directly. To be specific, given an anchor point located at $\boldsymbol{x}^a$ and its attributes $\mathcal{A}=\left\{f\in \mathbb{R}^{d},\boldsymbol{l}\in \mathbb{R}^6,\boldsymbol{o} \in \mathbb{R}^{3K}\right\}$, where $f$, $\boldsymbol{l}$, and $\boldsymbol{o}$ are the anchor feature, attribute scaling, and Gaussian offsets of anchor $\boldsymbol{x}_a$, respectively. $K$ is the number of Gaussian offsets belonging to anchor $\boldsymbol{x}_a$. During rendering, $f^{\bs{a}}$ are fed into MLPs to generate attributes $\mathcal{A}$ for Gaussians. The location of each Gaussian are calculated by adding $\bs{x}$ and Gaussian offsets $\bs{o}$ and $\bs{l}$ is applied to scale the location and shape of the Gaussians. 

Scaffold-GS achieves a similar rendering speed as the original 3DGS for the static scenes and the storage requirements are significantly reduced as Scaffold-GS only stores anchor points and MLPs for each scenes. The number of anchor points is much fewer than the number of Gaussians. As discussed earlier, the bottleneck of the rendering speed for dynamic scenes is the number of Gaussians. To this end, we extend the anchor-based strategy to the dynamic scene reconstruction for faster rendering speed.

\section{Method}

Our goal is to reconstruct a dynamic scene from a monocular video. We present the geometry and appearance of the dynamic scene with sparse anchor-grid initialization while modeling the deformation based on the sparse representation. The time-invariant attributes of Gaussians, \eg, color and opacity, are decoded by the tiny MLPs. Other attributes of the deformable objects, \eg, location, scale, and quaternion, vary across time, whereas those attributes of the static objects are consistent across time. We filter the anchors of the static objects with a tiny MLP and only the anchors of deformation objects are fed into corresponding MLPs for time-variant attributes. We formulate the movements of deformable Gaussians in a sparse semantic manner. Specifically, we query the movement of the deformed anchors at each time step and the movements of the assigned Gaussian offsets are calculated by the radial basis function (RBF) kernel. Please note that we do not need any additional 2D or 3D masks for supervision.

\subsection{Anchor-grid Initialization}
\label{sec:init}
The static anchor-grid representation is initialized by the sparse point cloud from COLMAP~\cite{schonberger2016colmap}. To be specific, given the 3D point cloud $\mathbf{P}\in \mathbb{R}^{M\times 3}$, we first voxelize the scene by: 
\begin{equation}
    \mathbf{A} = \left \lfloor \frac{\mathbf{P}}{\Delta d }  \right \rfloor \cdot \Delta d,
\end{equation}
where $\mathbf{A} = \left \{ \boldsymbol{a}_0,\boldsymbol{a}_1,\cdots,\boldsymbol{a}_{N-1} \right \}  \in \mathbb{R}^{N\times 3}$ is the structured anchor-grid, $\left \lfloor \cdot \right \rfloor $ is the floor operation, and $\Delta d$ is the voxel size. The floor operation removes the redundant and overdense points $\mathbf{P}$.

\subsection{Gaussian Attribute Derivation}

We first introduce how to derive Gaussian's attribute, \textit{i.e.}, opacity, quaternion, scaling, and color from anchor points at various timestep $t$. The attributes of each Gaussian are decoded from its feature and time positional encoding $\gamma$ by tiny MLPs $\varPhi_{*}(\cdot)$, where $\varPhi_{*}(\cdot)$ is the tiny MLP for the attribute $*$. Given one anchor $\boldsymbol{a}$, there are $K$ attributes decoded by $\varPhi_{*}(\cdot)$ where $K$ is the number of Gaussians belonging to the anchor $\boldsymbol{a}$. Time-invariant attributes, \textit{i.e.}, opacity and color, are decoded by $ \varPhi_{\alpha}( f_{\boldsymbol{a}})$ and $\varPhi_{c}(f_{\boldsymbol{a}} )$, where $f_{\boldsymbol{a}}$ are anchor's feature. Quaternion, scale, and Gaussian offsets vary across time. For anchor $\boldsymbol{a}$ with scale $\boldsymbol{s}_a$, the scale of its belonging $k^{th}$ Gaussian offsets at time $t$ is calculated by
\begin{equation}
    \boldsymbol{s}_{\boldsymbol{o}}^{k,t} = \boldsymbol{s}_{\boldsymbol{a}} +  \boldsymbol{s}_{\boldsymbol{o}}^{k} + \Delta \boldsymbol{s}_{\boldsymbol{o}}^{k,t},
\end{equation}
where $\boldsymbol{s}_{\boldsymbol{o}}^{k}\in \left \{ \boldsymbol{s}_{\boldsymbol{o}}^{0},\boldsymbol{s}_{\boldsymbol{o}}^{1},\cdots, \boldsymbol{s}_{\boldsymbol{o}}^{K-1} \right \} $ are $K$ learnable scales belonging to anchor $\boldsymbol{a}$ and $\Delta \boldsymbol{s}_{\boldsymbol{o}}^{k,t}$ is the offset variation at time $t$ and $\Delta \boldsymbol{x}_{\boldsymbol{o}}^{k,t}$ is calculated by
\begin{equation}
    \Delta \boldsymbol{s}_{\boldsymbol{o}}^{k,t} = \varPhi_{s}(\mathrm{cat}\left \{ f_{\boldsymbol{a}}, \gamma(t) \right \}),
\end{equation}
where $\mathrm{cat}$ is the concatenation operation. The quaternion $\boldsymbol{r}_{\boldsymbol{o}}^{k,t}$ is calculated in a similar manner.

\subsection{Gaussian Offsets Derivation with RBF Kernel}

As formulated in \cref{eq:deform}, previous methods~\cite{wu20234dgs,yang2023deformable,huang2023scgs} query Gaussians attributes by the deformation network $\mathcal{F}$. Intuitively, the movements of anchor $\boldsymbol{a}$'s $K$ Gaussians should be semantically aligned. For rigid objects, the deformation of $K$ Gaussians is exactly the same as its anchor $\boldsymbol{a}$'s movement. The strict assumption seriously affects the performance for rendering novel views of deformed objects because there are so many non-rigid objects in the rendering scenes. Another SOTA solution is to formulate the deformation of offsets based on anchors is k-nearest neighborhood method (KNN). For instance, SCGS~\cite{huang2023scgs} calculates the deformation of Gaussians based on $k$-nearest control points. However, KNN is calculated by the distance between points and the performance is degraded when the moving objects are separated after collision.

In order to formulate the deformation of objects in a semantic manner, we calculate the location of anchor $\boldsymbol{a}$'s $k^{th}$ Gaussian offsets at timestep $t$ with radial basis function (RBF) kernel. To be specific, we first apply a deformation MLP $\mathcal{F}$ to calculate the movement of anchor $\boldsymbol{a}$ across time $t$:
\begin{equation}
    \Delta \boldsymbol{x}_{\boldsymbol{a}}^t = \mathcal{F}(\gamma (\boldsymbol{x}_{\boldsymbol{a}}, t)),
\end{equation}
where $\boldsymbol{x}_{\boldsymbol{a}}$ is the location of anchor $\bs{a}$ and $\gamma$ is the position encoding of location and time. The movement of anchor $\bs{a}$'s $k^{th}$ Gaussians at time $t$ is calculated by
\begin{equation}
    \Delta \boldsymbol{x}_{\boldsymbol{o}}^{k,t} =  \mathcal{K}(f_{\boldsymbol{a}}, f_{\boldsymbol{o}}^k) \cdot \Delta \boldsymbol{x}_{\boldsymbol{a}}^t,
\end{equation}
where $f_{\boldsymbol{a}}$ and $f_{\boldsymbol{o}}^k$ are the features of anchor $\bs{a}$ and its $k^{th}$ belonging Gaussians.  $\mathcal{K}(\cdot, \cdot)$ is the kernel basis function on anchor's feature $f_{\boldsymbol{a}}$ and its offsets' feature $f_{\boldsymbol{o}}^k$. Here we use the common radial basis function (RBF) kernel: 
\begin{equation}
    \mathcal{K}(f_{\boldsymbol{a}}, f_{\boldsymbol{o}}^k) = \exp(-\frac{\left \| f_{\boldsymbol{a}} - f_{\boldsymbol{o}}^k \right \| ^2}{2\sigma ^2} ),
\end{equation}
where $\sigma$ is the covariance of the RBF kernel. We set $\sigma = 1$ in all experiments. Similar to calculating scale and quaternion, the location of anchor $\bs{a}$'s $k^{th}$ Gaussian is calculated as
\begin{equation}
    \boldsymbol{x}_{\boldsymbol{o}}^{k,t} = \boldsymbol{x}_{\boldsymbol{a}} +  \boldsymbol{x}_{\boldsymbol{o}}^{k} + \Delta \boldsymbol{x}_{\boldsymbol{o}}^{k,t},
\end{equation}

\subsection{Time Mask}

\begin {table*}[htbp]
\centering

\resizebox{\textwidth}{!}{
  \begin{tabular}{ccccccccccccccccc}
    \hline \multirow[b]{2}{*}{ Method } & \multicolumn{3}{c}{ Sieve } & \multicolumn{3}{c}{ Plate } & \multicolumn{3}{c}{ Bell } & \multicolumn{3}{c}{ Press } \\
                   & PSNR$\uparrow$             & SSIM$\uparrow$              & LPIPS$\downarrow$           & PSNR$\uparrow$             & SSIM$\uparrow$              & LPIPS$\downarrow$           & PSNR$\uparrow$             & SSIM$\uparrow$              & LPIPS$\downarrow$           & PSNR$\uparrow$             & SSIM $\uparrow$             & LPIPS$\downarrow$           \\

    \hline 3D-GS~\cite{kerbl20233d}   & 23.16                      & 0.8203                      & 0.2247                      & 16.14                      & 0.6970                      & 0.4093                      & 21.01                      & 0.7885                      & 0.2503                      & 22.89                      & 0.8163                      & 0.2904                      \\

    TiNeuVox~\cite{fang2022fast}       & 21.49                      & 0.8265                      & 0.3176                      & \cellcolor{second}20.58   & 0.8027  & 0.3317                      & 23.08                      & 0.8242  & 0.2568                      & 24.47                      & 0.8613                      & 0.3001                      \\

    HyperNeRF~\cite{park2021hypernerf}      & 25.43  & 0.8798 & 0.1645                      & 18.93                      & 0.7709                      & 0.2940  & 23.06                      & 0.8097                      & 0.2052                      & 26.15   & 0.8897   & 0.1959  \\

    NeRF-DS~\cite{yan2023nerfds}        & 25.78   & 0.8900   & 0.1472   & \cellcolor{third}20.54 & 0.8042 & \cellcolor{second}0.1996   & 23.19  & 0.8212                      & 0.1867  & 25.72                      & 0.8618  & 0.2047                      \\

    4DGS~\cite{wu20234dgs} & \cellcolor{second}26.11                      & \cellcolor{second}0.9193                      & \cellcolor{best}0.1107  & 20.41                      & \cellcolor{second}0.8311                      & \cellcolor{third}0.2010                      & 25.70 & \cellcolor{third}0.9088 & \cellcolor{best}0.1103 & \cellcolor{best}26.72  & \cellcolor{second}0.9031                      & \cellcolor{best}0.1301 \\

    SCGS~\cite{huang2023scgs} & \cellcolor{third}25.93                      & \cellcolor{third}0.9187                      & \cellcolor{third}0.1194  & 20.17                      & \cellcolor{third}0.8257                      & 0.2104                      & \cellcolor{second}25.97 & \cellcolor{second}0.9172 & \cellcolor{second}0.1167 & \cellcolor{third}26.57  & 0.8971                      & \cellcolor{third}0.1367 \\

    Deformable 3DGS~\cite{yang2023deformable}           & 25.70 & 0.8715  & 0.1504 & 20.48  & 0.8124   & 0.2224 & \cellcolor{third}25.74   & 0.8503   & 0.1537   & 26.01 & \cellcolor{third}0.8646 & 0.1905   \\
    
    Ours           & \cellcolor{best}27.12 & \cellcolor{best}0.9271  & \cellcolor{second}0.1151 & \cellcolor{best}21.21  & \cellcolor{best}0.8957   & \cellcolor{best}0.1873 & \cellcolor{best}26.01   & \cellcolor{best}0.9203   & \cellcolor{third}0.1204   & \cellcolor{second}26.61 & \cellcolor{best}0.9054 & \cellcolor{second}0.1313   \\

    \hline & \multicolumn{3}{c}{ Cup } & \multicolumn{3}{c}{ As } & \multicolumn{3}{c}{ Basin } & \multicolumn{3}{c}{ Mean } \\

    Method         & PSNR$\uparrow$             & SSIM$\uparrow$              & LPIPS$\downarrow$           & PSNR$\uparrow$             & SSIM$\uparrow$              & LPIPS$\downarrow$           & PSNR$\uparrow$             & SSIM$\uparrow$              & LPIPS$\downarrow$           & PSNR$\uparrow$             & SSIM $\uparrow$             & LPIPS$\downarrow$           \\

    \hline 3D-GS~\cite{kerbl20233d}   & 21.71                      & 0.8304                      & 0.2548                      & 22.69                      & 0.8017                      & 0.2994                      & 18.42                      & 0.7170                      & 0.3153                      & 20.29                      & 0.7816                      & 0.2920                      \\

    TiNeuVox~\cite{fang2022fast}       & 19.71                      & 0.8109                      & 0.3643                      & 21.26                      & 0.8289                      & 0.3967                      & \cellcolor{best}20.66   & 0.8145  & 0.2690                      & 21.61                      & 0.8234                      & 0.2766                      \\

    HyperNeRF~\cite{park2021hypernerf}      & 24.59                      & 0.8770  & 0.1650  & 25.58  & \cellcolor{second}0.8949   & 0.1777 & \cellcolor{second}20.41 & 0.8199   & 0.1911                      & 23.45                      & 0.8488  & 0.1990  \\

    NeRF-DS~\cite{yan2023nerfds}        & \cellcolor{second}24.91   & 0.8741                      & 0.1737                      & 25.13                      & 0.8778                      & 0.1741   & \cellcolor{third}19.96  & 0.8166 & 0.1855   & 23.60  & 0.8494 & 0.1816 \\

    4DGS~\cite{wu20234dgs} & 24.57  & \cellcolor{third}0.9102 & \cellcolor{best}0.1185 & \cellcolor{third}26.30 & \cellcolor{third}0.8917  & \cellcolor{third}0.1499                      & 19.01                      & \cellcolor{third}0.8277                      & \cellcolor{second}0.1631 & \cellcolor{second}24.18 & \cellcolor{third}0.8845                      &\cellcolor{best} 0.1405                      \\

    SCGS~\cite{huang2023scgs} & 24.32  & \cellcolor{second}0.9121 & \cellcolor{second}0.1207 & 26.17 & 0.8851  & \cellcolor{second}0.1491                      & 19.23                      & \cellcolor{best}0.8379                      & \cellcolor{best}0.1514 & 24.05 & \cellcolor{second}0.8848                      & \cellcolor{third}0.1439                      \\

    Deformable 3DGS~\cite{yang2023deformable}           & \cellcolor{third}24.86 & 0.8908   & 0.1532   & \cellcolor{second}26.31   & 0.8842 & 0.1783  & 19.67                      & 0.7934                      & 0.1901  & \cellcolor{third}24.11   & 0.8524   & 0.1769   \\
    Ours          & \cellcolor{best}25.08 & \cellcolor{best}0.9132   & \cellcolor{third}0.1225   & \cellcolor{best}26.65   & \cellcolor{best}0.9015 & \cellcolor{best}0.1472  & 19.91                      & \cellcolor{second}0.8351                      & \cellcolor{third}0.1640  & \cellcolor{best}24.65   & \cellcolor{best}0.8998   & \cellcolor{second}0.1411   \\
    \hline
  \end{tabular}}
  \caption{\textbf{Quantitative comparison on NeRF-DS dataset per-scene}. We color each cell as \colorbox{best}{best}, \colorbox{second}{second best}, and \colorbox{third}{third best}.}
\label{tab:nerfds-per}
\end{table*}

As discussed earlier, the scales, quaternion, and location of Gaussians belonging to static scenes are not changed over time. The rendering speed would be faster if we first filter the anchors belonging to static scenes and then only query time-variant attributes by MLPs for anchors that are responsible for deformation objects.
To this end, we introduce a time mask MLP $\varPhi_{mask}$ to filter the anchors responsible for the deformation objects. $\varPhi_{mask}$ is a tiny binary-classifier. All anchors' features $\boldsymbol{f}_{\boldsymbol{a}} = \left \{ f_{\boldsymbol{a}}^0,f_{\boldsymbol{a}}^1,\cdots,f_{\boldsymbol{a}}^{N-1} \right \}  \in \mathbb{R}^{N\times d}$ are fed into $\varPhi_{mask}$. Only the anchors with output label 1 are further fed into corresponding MLPs for time-variant attributes. We further introduce a time mask regularization term $\mathcal{L}_{t-mask}$ to optimize all the learnable parameters with few anchors that are responsible for deformation objects:
\begin{equation}
    \mathcal{L}_{t-mask} = \sum_{n=0}^{N-1} \varPhi_{mask}(f_{\boldsymbol{a}}^{n}) / N,
\end{equation}
where $N$ is the number of anchors.

\subsection{Densify and Pruning for Anchors}

Though applying sparse anchor points improves the rendering speed greatly, we still need to densify the 3D Gaussians for higher rendering performance. Similar to \cite{lu2023scaffold, kerbl20233d}, we compute the gradients of the anchor points over 100 iterations and gradients greater than threshold $\epsilon$ will be densified. The voxel of the densified anchor points are divided into $m$ sub-voxels and the anchor points are placed to each sub-voxel. The anchor feature of each sub-voxel are copied from the densified anchor and $K$ offsets are further initialized for the anchor of each sub-voxel. For pruning redundant anchor points, we calculate the average opacity of the Gaussian offsets for each anchor point. The anchor is removed if the average opacity of its belonging Gaussian offsets is lower than the pre-defined threshold.

\subsection{Loss Function}

Similar to previous methods~\cite{kerbl20233d,huang2023scgs,yang2023deformable,wu20234dgs}, we optimize the feature $f$ each anchor $\boldsymbol{a}$ and its offsets $\boldsymbol{o}_{\boldsymbol{a}}$, tiny MLPs $F_{\alpha}$, $F_{r}$,  $F_{s}$, and $F_{c}$ for the attributes of Gaussians, and anchor's deformable network with respect to the $\mathcal{L}_1$ loss and SSIM loss $\mathcal{L}_{SSIM}$ with a time mask regularization term $\mathcal{L}_{t-mask}$ over the rendered RBG images to supervise the training process. Our final loss function is defined as:
\begin{equation}
    \mathcal{L} = (1-\lambda)\mathcal{L}_1 + \lambda \mathcal{L}_{SSIM} + \lambda_t \mathcal{L}_{t-mask},
\end{equation}
where $\lambda = 0.2$ and $\lambda_t = 0.2$ in our all experiments. It should be noted that during the training process, the position of each anchor $\boldsymbol{a} \in \mathbf{A}$ remains frozen and does not undergo any updates.

\vspace{-0.4cm}
\begin{table}[htbp]
	\centering
\fontsize{7}{9}\selectfont
        
	\begin{tabular}{c|cc|c} 
		\hline  
		Model                                   & PSNR(dB)↑            & MS-SSIM↑              & FPS↑               \\ 
		\hline  
		Nerfies~\cite{park2021nerfies}          & 22.2                   & 0.803                   & $<$ 1                \\
		HyperNeRF~\cite{park2021hypernerf}      & 22.4                   & 0.814                   & $<$ 1                \\
		TiNeuVox-B~\cite{fang2022fast}              & 24.3 & 0.836                   & 1                    \\
		3D-GS~\cite{kerbl20233d}                       & 19.7                   & 0.680                   & \cellcolor{third}32   \\
		FFDNeRF~\cite{guo2023forwardflowfornvs} & 24.2                   & \cellcolor{third}0.842 & 0.05                 \\
        Deformable 3DGS~\cite{yang2023deformable}                                   & \cellcolor{third}25.0   & 0.822   & 13                   \\
        4DGS~\cite{wu20234dgs}                                    & \cellcolor{second}25.2   & \cellcolor{second}0.845   & \cellcolor{second}34  \\
        SCGS~\cite{huang2023scgs}                                    & 24.6   & 0.813   & 12                   \\

		Ours                                    & \cellcolor{best}25.7   & \cellcolor{best}0.860   & \cellcolor{best}117  \\
		\hline
	\end{tabular}  
 \caption{\textbf{Quantitative results on HyperNeRF's~\cite{park2021hypernerf} vrig dataset.} The rendering resolution is set to 960$\times$540. We color each cell as \colorbox{best}{best}, \colorbox{second}{second best}, and \colorbox{third}{third best}.}
	\label{tab:hypernerf}
\end{table}
\vspace{-0.6cm}

\section{Experiments}

\subsection{Experimental Setup}

\paragraph{Dataset and metrics.} 

We conducted extensive experiments on two real-world datasets: the HyperNeRF dataset~\cite{park2021hypernerf} and the NeRF-DS dataset~\cite{yan2023nerfds}. The division on the training and testing subsets and other experimental protocols are perfectly aligned with the original papers. The metrics we apply to evaluate the performance are Peak Signal-to-Noise Ratio (RSNR), Structural Similarity (SSIM), and Learned Perceptual Image Patch Similarity (LPIPS)~\cite{zhang2018unreasonable}. Apart from these commonly used metrics, we additionally report the training time and the rendering speed (FPS) for model compactness and efficiency. We report the metrics per scene on the NeRF-DS dataset~\cite{yan2023nerfds} and the averaged metrics over all scenes on the HyperNeRF~\cite{park2021hypernerf} dataset.

\begin{figure*}[htbp]
    \centering 
    \includegraphics[width=0.99\textwidth]{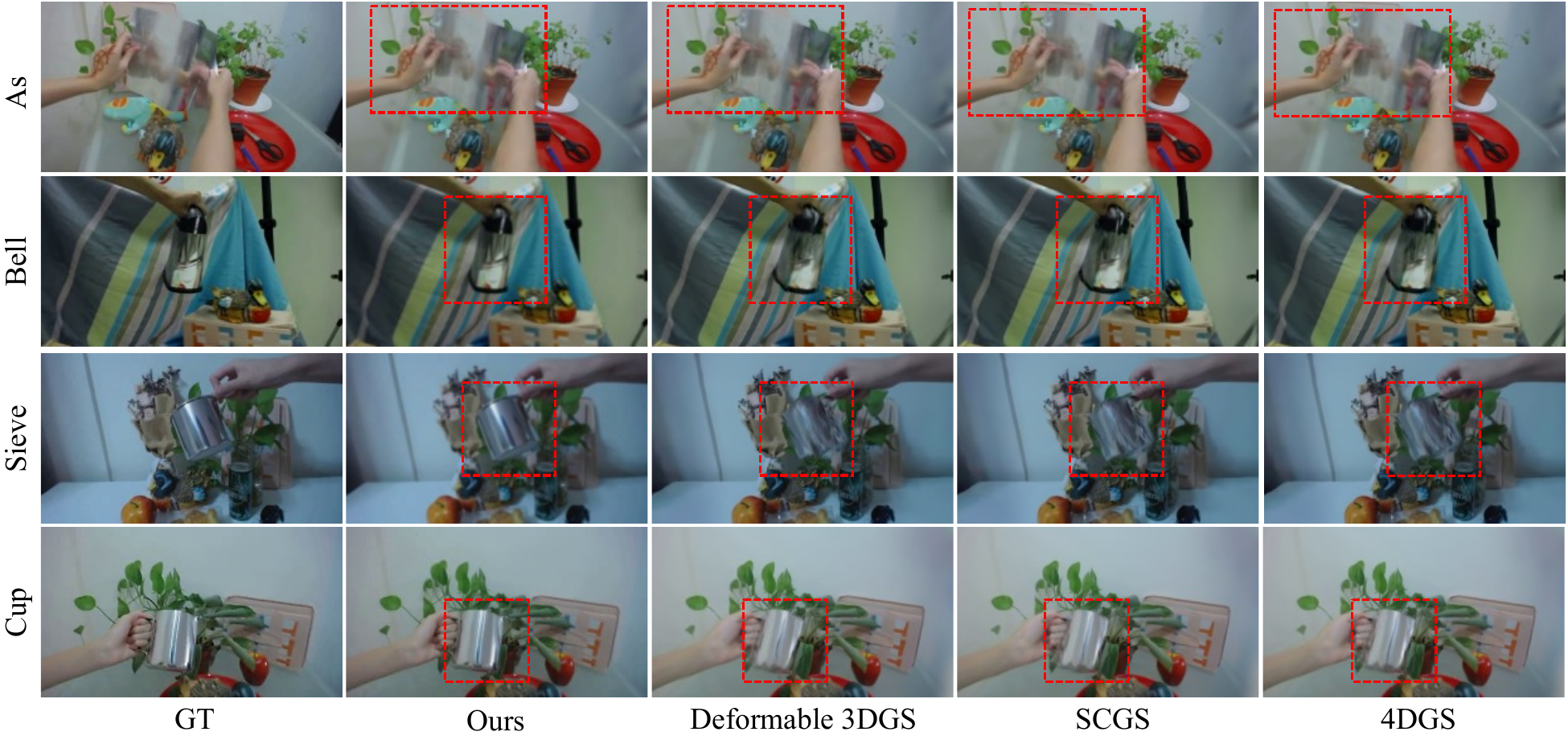}
    \caption{\textbf{Qualitative comparison on the NeRF-DS dataset~\cite{yan2023nerfds}}. Compared with other SOTA methods, our method reconstructs finer details and produces a structured rendering of the moving objects, \textit{e.g.}, the cup on human's hand.} 
    \label{fig:nerfds}
\end{figure*}

\paragraph{Baselines and implementation.}
To evaluate the performance of novel view synthesis for real-world dynamic scenes, we conducted benchmarks against several state-of-the-art methods in the field, including NeRF-based methods~\cite{fang2022fast,park2021hypernerf,yan2023nerfds,park2021nerfies,song2023nerfplayer,attal2023hyperreel,kplanes,lin2023im4d,msth} and 3DGS-based methods~\cite{kerbl20233d,huang2023scgs,yang2023deformable,wu20234dgs}. Our implementation is primarily based on PyTroch~\cite{paszke2019pytorch} framework and evaluated by a Nvidia V100 GPU. Most of our hyper-parameters follow 3DGS~\cite{kerbl20233d}. Our pipeline is trained for 30k iterations. We set $k=10$ offset Gaussians for each anchor and the dimension of the anchor features $f_{\boldsymbol{a}}$ and offset features $f_{\boldsymbol{o}}$ is 8. The tiny MLPs $\varPhi_*$ for Gaussian attributes are two-layer MLPs and the dimension of the middle layer is 64. The anchor's deformation MLP is four-layer fully connected layers that employ ReLU activation and the dimension of intermediate layers is 128. The voxel size for the initialization of anchors is 0.6.

\subsection{Quantitative Comparisons}
\label{sec:quan}

\paragraph{NeRF-DS dataset.}

We first compare our method with baselines using monocular real-world NeRF-DS~\cite{yan2023nerfds} dataset. The experimental results on the NeRF-DS dataset, presented in \cref{tab:nerfds-per}, clearly demonstrate the superiority of our proposed method across multiple evaluation metrics and scenes. our method achieves the highest mean PSNR (24.39) and SSIM (0.8873) across all scenes, indicating its superior performance and consistency. Moreover, with the help of the structured anchor and the formulation between anchor and Gaussian offsets, our method achieves high rendering performance for moving objects. Please refer to \cref{fig:nerfds} for visualization. 

\begin{figure}[h]
    \centering 
    \includegraphics[width=0.45\textwidth]{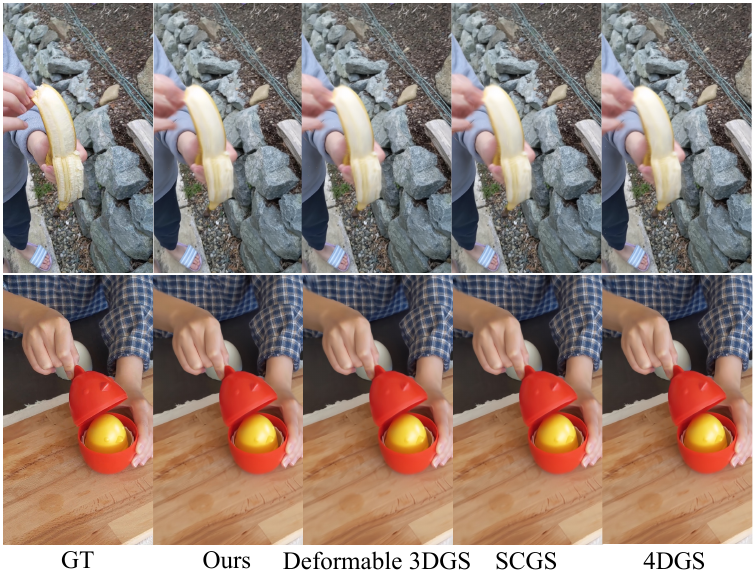}
    \caption{\textbf{Qualitative comparison on the HyperNeRF dataset~\cite{park2021hypernerf}.} Our EDGS reconstructs detailed texture and reliable structure compared with other SOTA methods.} 
    \vspace{-0.6cm}
    \label{fig:hypernerf}
\end{figure}

\paragraph{HyperNeRF dataset.}

We compare our method with other baselines on the HyperNeRF dataset~\cite{park2021hypernerf}. The experimental results on HyperNeRF's VRIG dataset, shown in \cref{tab:hypernerf}, highlight the superior performance of our proposed method. Achieving the highest PSNR of 25.7 dB and MS-SSIM of 0.860, our approach ensures exceptional image quality and structural fidelity. Furthermore, it demonstrates remarkable efficiency with a rendering time of just 20 minutes and an impressive 117 FPS, significantly outperforming other methods. Additionally, our method requires only 7K Gaussians which need to query time-variant attributes by MLP, the lowest among all compared techniques, underscoring its computational efficiency. These results validate the robustness and practicality of our EDGS for dynamic scene reconstruction.

\begin{figure*}[h]
    \centering 
    \includegraphics[width=0.95\textwidth]{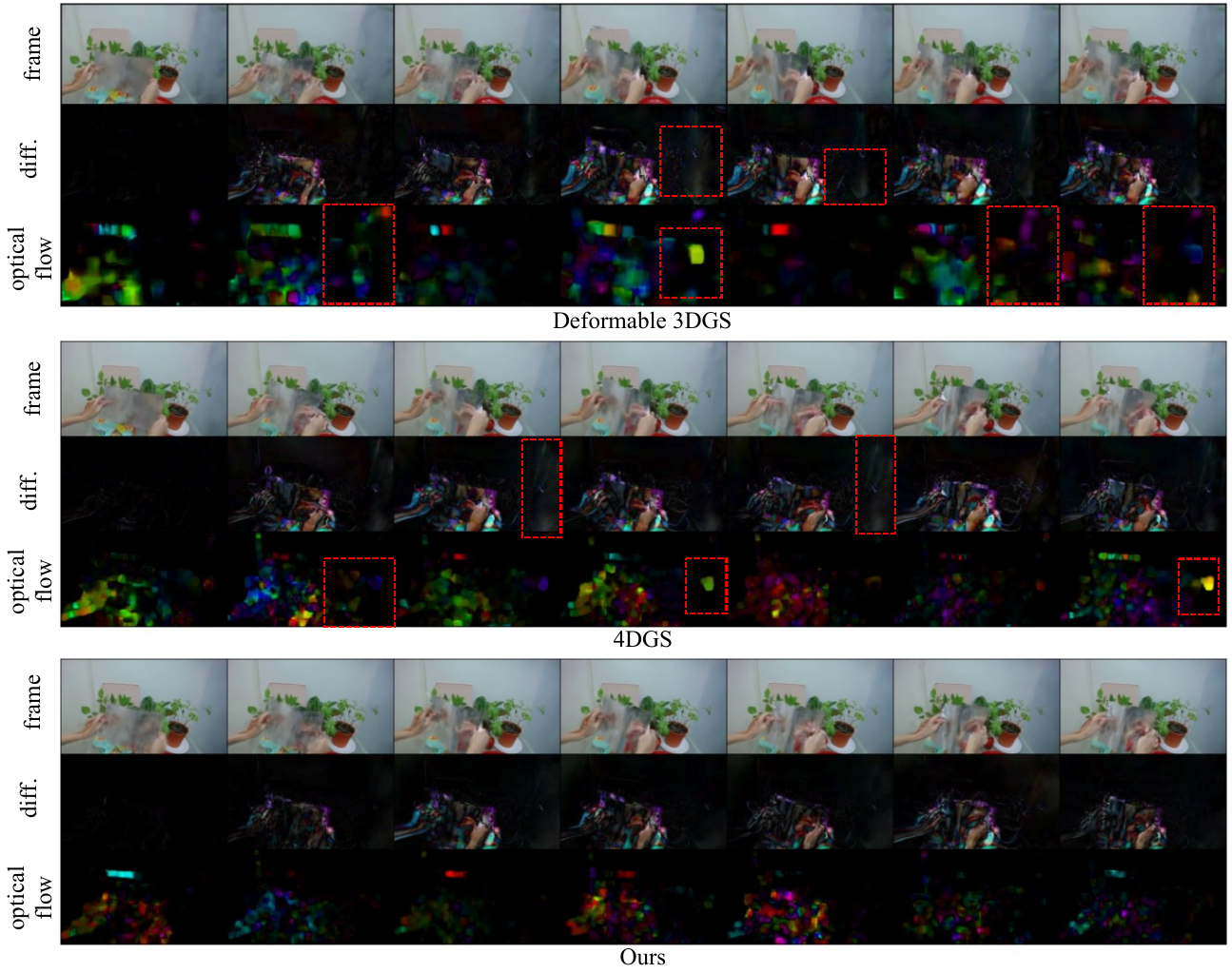}
    \caption{\textbf{Visuazization of the difference map (diff.) and the optical flow with fixed camera views.} We synthesis fixed novel view across time for \cite{yang2023deformable,wu20234dgs} and ours. The $1^{st}$ row is the rendered frames at various time steps. The $2^{nd}$ and $3^{rd}$ rows are the difference map between $t^{th}$ frame and the $1^{st}$ frame and the optical flow, respectively. The response in the highlighted red area indicates that the static area rendered by deformable GS and 4DGS is jittering. Our method achieves better quality for static and dynamic objects.} 
    \label{fig:error}
\end{figure*} 

\subsection{Qualitative Comparisons}
\label{sec:qua}

We conduct qualitative comparisons to illustrate the advantages of our method over other SOTA methods. The comparisons on the NeRF-DS dataset are shown in \cref{fig:nerfds}. Please zoom in to the highlighted red region for rendering comparisons of the moving objects. Compared with other SOTA methods, our method reconstructs finer details and produces a more structured rendering of the moving objects, \textit{e.g.}, the cup in hand. 

We also visualize the rendered results with fixed cameras and show the difference map between each frame and the $1^{st}$ frame and the optical flow of the rendered video in \cref{fig:error}. As highlighted in the red box in \cref{fig:error}, deformable 3DGS and 4DGS fail to render the static area. The difference map and the optical flow at the static area indicates that the static area is jittering for those methods. We render the static and dynamic objects more accurate with fewer jittering issues because of the time-mask MLP. 

We also visualize the mask predicted by the time-mask MLP in \cref{fig:feat}. The visualization comparisons on the HyperNeRF dataset~\cite{park2021hypernerf} are shown in \cref{fig:hypernerf}. Compared to other SOTA methods, our EDGS reconstructs finer details (\eg, the red chicken toy and the banana in hand) and produces a more structured rendering of moving objects.

\subsection{Ablation Studies}

\begin{figure}[htbp]
\centering 
\includegraphics[width=0.41\textwidth]{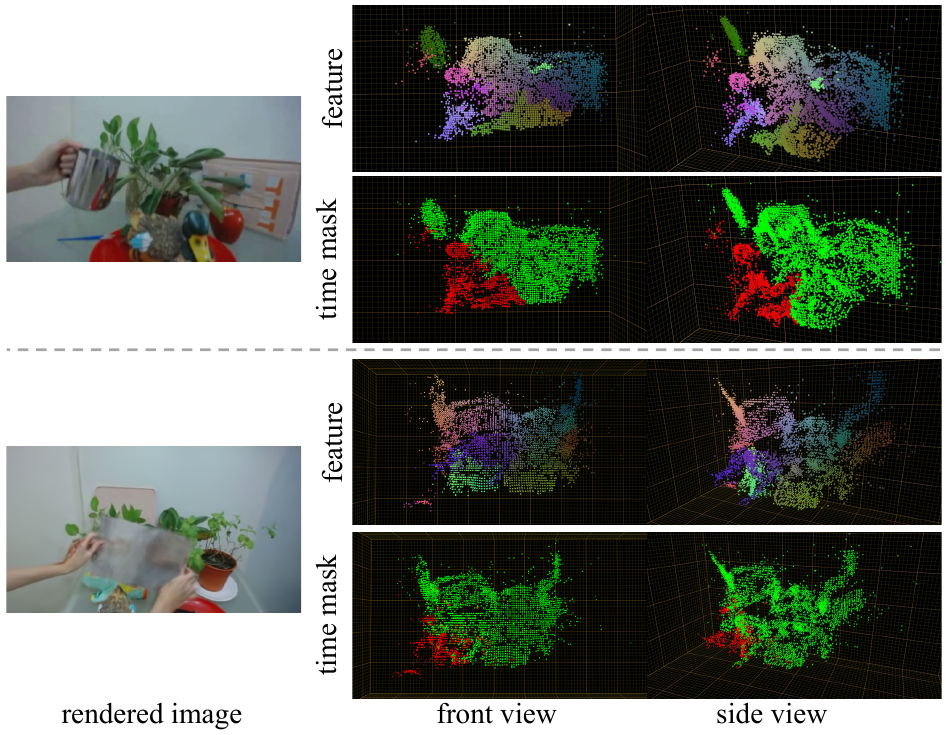}
\caption{\textbf{Visualization of anchor features and time mask.} We visualize the rendered images, anchor features, and time masks for two deformation scenes. Anchor features are visualized using the UMAP function. In the time-mask visualization, the time masks are predicted by time-mask MLP $\varPhi_{mask}$. \textcolor{green}{Green} anchors represent static scenes, while \textcolor{red}{red} anchors indicate deformation scenes.} 
\label{fig:feat}
\end{figure}

\paragraph{Efficacy of anchor-voxel strategy and time-mask MLP.} 
We assess the efficiency of the anchor-voxel strategy and time-mask MLP on three scenes from the NeRF-DS~\cite{yan2023nerfds} dataset. Without the anchor-voxel strategy, rendering speed drops to around 20 FPS due to redundant densified Gaussians, showing that the bottleneck in rendering deformation scenes is the number of Gaussians. Introducing the time-mask MLP further improves rendering speed by filtering out anchors corresponding to static areas, allowing only those in dynamic areas to be queried for time-variant attributes. This also enhances rendering quality (\textit{e.g.} PSNR increases from 26.18 to 27.12 on the sieve scene), likely because the MLPs are optimized with fewer anchor points, making the optimization process more effective. See \cref{fig:feat} for the time mask visualization.

\begin{table}[h]
    \centering
    
    \resizebox{0.4\textwidth}{!}{
      \begin{tabular}{c|cc|cc|cc}
      \toprule
      \multirow{2}[2]{*}{scene} & \multicolumn{2}{c|}{As} & \multicolumn{2}{c|}{Bell} & \multicolumn{2}{c}{Sieve} \\
            & \multicolumn{1}{l}{PSNR} & \multicolumn{1}{l|}{FPS} & \multicolumn{1}{l}{PSNR} & \multicolumn{1}{l|}{FPS} & \multicolumn{1}{l}{PSNR} & \multicolumn{1}{l}{FPS} \\
      \midrule
      w/o anchor-grid init & 24.57 & 27    & 24.31 & 23    & 24.74 & 17 \\
      w/o time-mask & 26.02 & 127   & 25.78 & 132   & 26.18 & 122 \\
      \midrule
      full model & 26.65 & 152   & 26.01 & 147   & 27.12 & 151 \\
      \bottomrule
      \end{tabular}%
      }
      \caption{\textbf{Effects of anchor-voxel strategy and time-mask MLP.} The anchor-grid strategy is crucial for rendering speed due to the reduced number of Gaussians. The time-mask MLP filters time-variant anchors, further reducing the number of anchors required for time-variant attribute queries.}
    \label{tab:anchor}%
  \end{table}%

\paragraph{Efficacy of time-variant Gaussian offsets derivation with various strategies.} 
\cref{tab:kernel} compares time-variant Gaussian offsets derived using different strategies across three scenes on the NeRF-DS~\cite{yan2023nerfds} dataset. Regarding the rigid transformation strategy, we assume that the movements of the anchors and their corresponding Gaussians same. The lowest performance shows that simply regarding all deformation objects as rigid ones degrades the performance. We further calculate the similarity between the feature of anchor and Gaussian offsets using KNN method and cosine similarity. The KNN strategy performs well for the reconstruction of single deformation objects~\cite{huang2023scgs} and the performance degrades on the real-world scenes. Cosine similarity achieves similar performance compared with RBF kernel but is VRAM-consuming because the similarity is calculated by a $N \times (KN)$ huge matrix.

\begin{table}[htbp]
  \centering
  
  \resizebox{0.45\textwidth}{!}{
    \begin{tabular}{c|cc|cc|cc}
    \toprule
    \multirow{2}[1]{*}{scene} & \multicolumn{2}{c|}{As} & \multicolumn{2}{c|}{Bell} & \multicolumn{2}{c}{Sieve} \\
          & PSNR  & SSIM  & PSNR  & SSIM  & PSNR  & SSIM \\
    \midrule
    rigid transformation & 23.51 & 0.8259 & 24.21 & 0.8571 & 23.95 & 0.8750 \\
    KNN   & 25.75 & 0.8954 & 25.53 & 0.8816 & 25.68 & 0.8907 \\
    cosine similarity & 26.31 & 0.8928 & 25.74 & 0.8807 & 26.57 & 0.9155 \\
    \midrule
    RBF kernel & 26.65 & 0.9015 & 26.01 & 0.9203 & 27.12 & 0.9271 \\
    \bottomrule
    \end{tabular}%
    }
    \caption{\textbf{Gaussian offsets derivation with various strategies.} We conduct four different strategies for formulating the variation of the Gaussian offsets across time.}
  \label{tab:kernel}%
\end{table}%

\paragraph{Visualization of anchor features and time mask.} We perform an analysis of the learnable anchor features to assess their effectiveness in dynamic scene reconstruction. As shown in \cref{fig:feat}, the clustered patterns indicate that the anchor features learn and encode similar semantic meanings across both static and dynamic objects of the scene.  Additionally, the time-mask MLP $\varPhi_{mask}$ effectively classifies time-variant anchors without any mask supervision, demonstrating its ability to adaptively distinguish between static and dynamic regions.

\section{Conclusion}
In conclusion, we develop Efficient Dynamic Gaussian Splatting (EDGS). It efficiently models dynamic scenes by emphasizing sparse, time-variant attributes and selectively processing static objects. This strategy significantly reduces computational complexity while preserving high rendering quality. Additionally, the incorporation of a classical kernel for motion flow optimization further enhances this process. Our evaluations using the NeRF-DS and HyperNeRF datasets show that EDGS not only achieves faster rendering speeds but also higher Peak Signal-to-Noise Ratio (PSNR) scores, surpassing current state-of-the-art methods.

\section*{Acknowledgements} 
This project is supported by the Ministry of Education, Singapore, under its Academic Research Fund Tier 2 (Award Number: MOE-T2EP20122-0006), and the National Research Foundation, Singapore, under its Medium Sized Center for Advanced Robotics Technology Innovation.

\bibliography{aaai25}

\end{document}